\def\year{2018}\relax
\documentclass[letterpaper]{article} 
\usepackage{aaai18}  
\usepackage{times}  
\usepackage{helvet}  
\usepackage{courier}  
\usepackage{url}  
\usepackage{graphicx}  
\usepackage{mathrsfs}
\usepackage{multirow}
\usepackage{epsfig}
\usepackage{graphicx,amsmath,subfigure, url }
\newcommand{\stitle}[1]{{\bf #1}}

\newcommand{\eat}[1]{}

\eat{ATRank: An Attention-Based Heterogeneous User Behavior Modeling \\ Framework for Recommendation}

\begin{document}
%
\title{ATRank: An Attention-Based User Behavior Modeling Framework for Recommendation}

\author{
Chang Zhou\textsuperscript{1},
Jinze Bai\textsuperscript{2},
Junshuai Song\textsuperscript{2},
Xiaofei Liu\textsuperscript{1},
Zhengchao Zhao\textsuperscript{1},
Xiusi Chen\textsuperscript{2},
Jun Gao\textsuperscript{2}\\
\textsuperscript{1}Alibaba Group\\
\textsuperscript{2}Key Laboratory of High Confidence Software Technologies, EECS, Peking University\\
\{ericzhou.zc, hsiaofei.hfl, buqing\}@alibaba-inc.com,
\{baijinze, songjs, xiusichen, gaojun\}@pku.edu.cn
}
\maketitle
\begin{abstract}
\eat{For any system that receives only the implicit feedback from users, it's usually hard to model its users.}
A user can be represented as what he/she does along the history. A common way to deal with the user modeling problem is to manually extract all kinds of aggregated features over the heterogeneous behaviors, 
which may fail to fully represent the data itself due to limited human instinct. Recent works usually use RNN-based methods to give an overall embedding of a behavior sequence, which then could be exploited by the downstream applications. However, this can only preserve very limited information, or aggregated memories of a person. When a downstream application requires to facilitate the modeled user features, it may lose the integrity of the specific highly correlated behavior of the user, and introduce noises derived from unrelated behaviors. This paper proposes an attention based user behavior modeling framework called ATRank, which we mainly use for recommendation tasks. Heterogeneous user behaviors are considered in our model that we project all types of behaviors into multiple latent semantic spaces, where influence can be made among the behaviors via self-attention. Downstream applications then can use the user behavior vectors via vanilla attention. Experiments show that ATRank can achieve better performance and faster training process. We further explore ATRank to use one unified model to predict different types of user behaviors at the same time, showing a comparable performance with the highly optimized individual models.

\end{abstract}

\section{Introduction}

\eat{
As a word can be represented by the surrounding context \cite{w2v}, a user can be represented by his/her behaviors along the history.
The information of various kinds of user behaviors that could be collected today is much more abundant than ever, which makes it increasingly important to utilize those 
user behaviors to provide better personalized services. \eat{to its users if it's capable to model their behaviors in different aspects.}
\eat{ of the user information are collected.}
}

As a word can be represented by the surrounding context \cite{w2v}, a user can be represented by his/her behaviors along the history.
For downstream tasks like ranking in recommendation system, traditional ways to represent a user are to extract all kinds of hand-crafted features aggregated over different types of user behaviors. This feature engineering procedure guided by human instinct may fail to fully represent the data itself, and it requires too much work. 
Besides, the aggregated features also lose information of any individual behavior that could be precisely related with the object that needs to be predicted in the downstream application.

As the user behaviors naturally form a sequence over the timeline, RNN/CNN structures are usually exploited to encode the behavior sequence for downstream applications as in the encoder-decoder framework.
However, for RNN based methods, long-term dependencies are still very hard to preserve even using the advanced memory cell structures like LSTM and GRU \cite{lstm,gru}. 
RNN also has a disadvantage that both the offline training and the online prediction process are time-consuming, due to its recursive natural which is hard to parallelize. 
Recently, it's shown that CNN-based encoding methods can also achieve comparable performance with RNN in many sequence processing tasks \cite{cnn_nlp}. Though it's highly parallelizable, the longest length of the interaction paths between any two positions in the CNN network is $log_k(n)$, where $n$ is the number of user behaviors and $k$ is the kernel width.

Both the basic RNN and CNN encoders suffer from the problem that the fixed-size encoding vector may not support both short and long sequences well. 
\eat{And for the recommendation task, the fixed-length encoding is an overall aggregated user status, which is then }
The attention mechanism is then introduced to provide the ability to reference specific records dynamically in the decoder, 
which has already achieved great successes in fields like Machine Translation, Image Caption in recent years.
Downstream applications of behavior modeling like recommendation can also utilize the attention mechanism, since the item to be ranked may only be related to very small parts of the user behaviors. 

However, we show that the one-dimensional attention score between any two vectors may neutralize their relationships in different semantic spaces.
The attention-based pooling can be formularized as $C=\sum\limits_{i=1}^{n}{a_i \vec{v_i}}$,
where $a_i$ is a scalar. We can see that, each element in $\vec{v_i}$ will multiply by the same $a_i$, such that different semantics of $\vec{v_i}$ can only compromise to use this overall multiplier, 
which makes it hard to preserve only the highly related semantics while discard those unrelated parts in the weighted average vector.

\eat{
More importantly,  a behavior done by a person $u$ can be represented statically that the same input could probably refer to the same encoding output, but the actual meaning of this behavior to $u$ could be affected by other behaviors performed by the same person. \eat{The attention of a behavior for $u$ may be distracted or enhanced.}
Let's consider a common case where the attention distraction or enhancement of a behavior takes place.
A user has browsed a lot of cellphones, which may reveal that he wants to buy a phone. Later, he orders one of the phones. After that, the intents that he wants to buy a phone may be degraded,  or the attention for phones is distracted, which probably resulting in that the strength of the phone browsing behavior may be weaken. However, the attention for complementary items may be lifted up. 

These interactions among the behaviors performed by the same person can be modeled using self-attention, which is fast to train, predict and has a nice interpretation about what's going on in the process of the downstream applications. 
}

Another important issue is that the user behaviors are naturally heterogeneous, highly flexible, and thus hard to model.
For large companies today, various kinds of user behaviors could be collected, which makes it increasingly important to utilize those 
user behaviors to provide better personalized services. 
Take the e-business recommendation service as an example, a user may browse/buy/mark items, receive/use coupons, click ads, search keywords, write down reviews, or even watch videos and live shows
offered by a shop, each of which reveals some aspects of the user that would be helpful to build more comprehensive user models, providing a better understanding of the user intents.
Though heterogeneous data representation can be learnt by minimizing the distances in the projected latent spaces \cite{transE,transR,hetero_emb}, there are no explicit supervisions like mapping or inferencing between any pair of user behaviors that could help build the individual behavior representations.

In this paper, we propose an attention-based user behavior modeling framework called ATRank, which we currently use for recommendation tasks. 
We first project variable-length heterogenous behavior representations into multiple latent spaces, where behavior interactions can be made under common semantics. Then we exploit the power of self-attention mechanism to model each user behavior after considering the influences brought by other behaviors. 
We perform vanilla attention between these attention vectors and the ranking item vector,  whose outputs are fed into a ranking neural network.  
It's observed that self-attention with time encoding can be a replacement for complex RNN and CNN structures in the sequential behavior encoding, which is fast in both training and prediction phase.  Experiments also show that ATRank achieves better performance while is faster in training.
We further explore the model to predict multiple types of user behaviors at the same time using only one unified model, showing a comparable performance with the highly optimized individual models.

\section{Related Works}

\eat{
Recent works usually use RNN-based methods to capture the aggregated features of the sequential behaviors automatically. The real valued temporal information is exploited by \cite{strnn} along with the discrete positions along the sequence.
}
 \eat{methods to evaluate similarities of different entities w.r.t different edge types. }
 
\stitle{Context Aware Recommendation.} An industrial recommendation system usually has varieties of models to extract different aspects of the user features, e.g., user gender, income, affordance, categorial preference, etc, all from the user behaviors, trying to capture both long-term and short-term user interests. 
 Then it builds different models for each recommendation scenario using those extracted features, either continuous or categorial \cite{youtube,wide_n_deep}. 
\eat{Sometimes it also introduces factors like fatigue using exhausted feature engineering to prevent from seeing too many items of the same tastes.} These are multi-step tasks and it's hard to optimize jointly. 

RNN based methods are studied for recommendation in academic fields in recent years, which builds the recommendation context directly from the behaviors for each user \cite{context_aware_fast,context_aware_music,rnn_163}. Though it's an more elegant way to encode user context, it still suffers from several difficulties. First, RNN is hard to parallelize in prediction phase, which makes it not easy to ensure the response time to be low enough for a commercial recommendation system. Second, the RNN embedding of the user behaviors is a fix-sized, aggregated status, which is not well suited for modeling both long and short behavior sequences. It can easily fail to preserve specific behavior information when enrolled in the downstream applications.  

\eat{
When  recommendation task , or the sequential prediction tasks \cite{}.
An end to end model is highly needed: what if we build user models directly from the raw feature of the sequential user behaviors.  
}

\eat{The basic idea of attention is that instead of attempting to learn a single vector representation for each sentence, we instead keep around vectors for every word in the input sentence, and reference these vectors at each decoding step.
}
\stitle{Attention and Self-Attention. }
Attention is introduced by \cite{attention_raw} firstly in the encoder-decoder framework, to provide more accurate alignment for each position in the machine translation task. 
Instead of preserving only one single vector representation for the whole object in the encoder, it keeps the vectors for each element as well, so that the decoder can reference these vectors at any decoding step.
Recently, attention based methods are widely applied in many other tasks like reading comprehension \cite{self_attention_read,attention_read,self_attention}, ads recommendation \cite{attention_ad_ali,attention_ad}, computer vision \cite{cv_attention}, etc. 

Self-attentions are also studied in different mechanisms \cite{google_attention,self_attention,self_attention_read}, in which inner-relations of the data at the encoder side are considered.
Note that, both work of \cite{google_attention,self_attention} show that project each word onto multiple spaces could improve the performance of their own tasks.

\eat{\cite{self_attention} captures the attention score calculated by the importance of each word in the projected space, and it interpretates this phenomenon as that different subspaces may reveal different semantics. }

\eat{A user can be modeled as aggregated features extracted manually from each behavior group \cite{hydra}, while those behaviors have different attributes.}
\stitle{Heterogeneous Behavior Modeling.} 
Heterogeneous data representation is heavily studied in domains like knowledge graph completion and multimodal learning. 
In the domain of knowledge graph completion, lots of works have been proposed to learn heterogeneous entity and relation representations by minimizing the distances of the linear projected entities in the relation-type semantic subspace \cite{transE,transR,transH}. Similar idea is adopted in multimodal learning tasks like image caption \cite{image_caption} and 
audio-visual speech classification \cite{visual_speech}. 

\eat{Individual behavior modeling suffers from the problem that it has no explicit supervision about the relationships between any two behaviors. The only supervision comes from the inner-relation between the user behavior history and the next behavior it takes.
In this paper, we share similar ideas by projecting each behavior embedding into multiple semantic subspaces, considering that each subspace may convey different aspect of the heterogeneous behavior information.
}

\eat{

Vanilla attention is 
It preserves the sequence integrity until the generation/ranking step, so that downstream tasks can exploit the full power of the sequence.

2. Good interpretability  

Why We need self-attention?

1. Self-attention is introduced by \cite{attention_raw}??, and is further studied by \cite{google_attention,self_attention,self_attention_read} [CNN, corpus]. 

2. Google, Reading
Both works of \cite{google_attention,self_attention} show that project each word into multiple subspaces could improve the performance, and \cite{self_attention} interpreted this phenomenon as different subspaces may reveal different semantics. 
}
\section{Self-Attention Based Behavior Modeling Framework}

\eat{
\begin{figure}[htp]
\centering
\subfigure[Directed]{
       \psfig{figure=fig/star_directed.png,width=2.8 in,height=1 in}
         \label{Figure:d}
      }
\caption{Asymmetric Proximity in both Directed and Undirected Graphs. Intuitively, $Sim(A,C)$ is not equal with $Sim(C,A)$ in both cases, due to their asymmetric local structures. Even for undirected graphs, the learning model should treat the two sampled paths $C\rightarrow B\rightarrow A$ and $A\rightarrow B\rightarrow C$ differently as directed sequences, since the probabilities of this two sampled paths are quite different, which means the occurrence number of $(C,A)$ and $(A,C)$ in a graph could vary a lot in many real world applications that consider the asymmetric relations. }
\label{Figure:star}
\end{figure}
}

\begin{figure*}
\centering
\includegraphics [width=5.2 in, height=2.8 in]{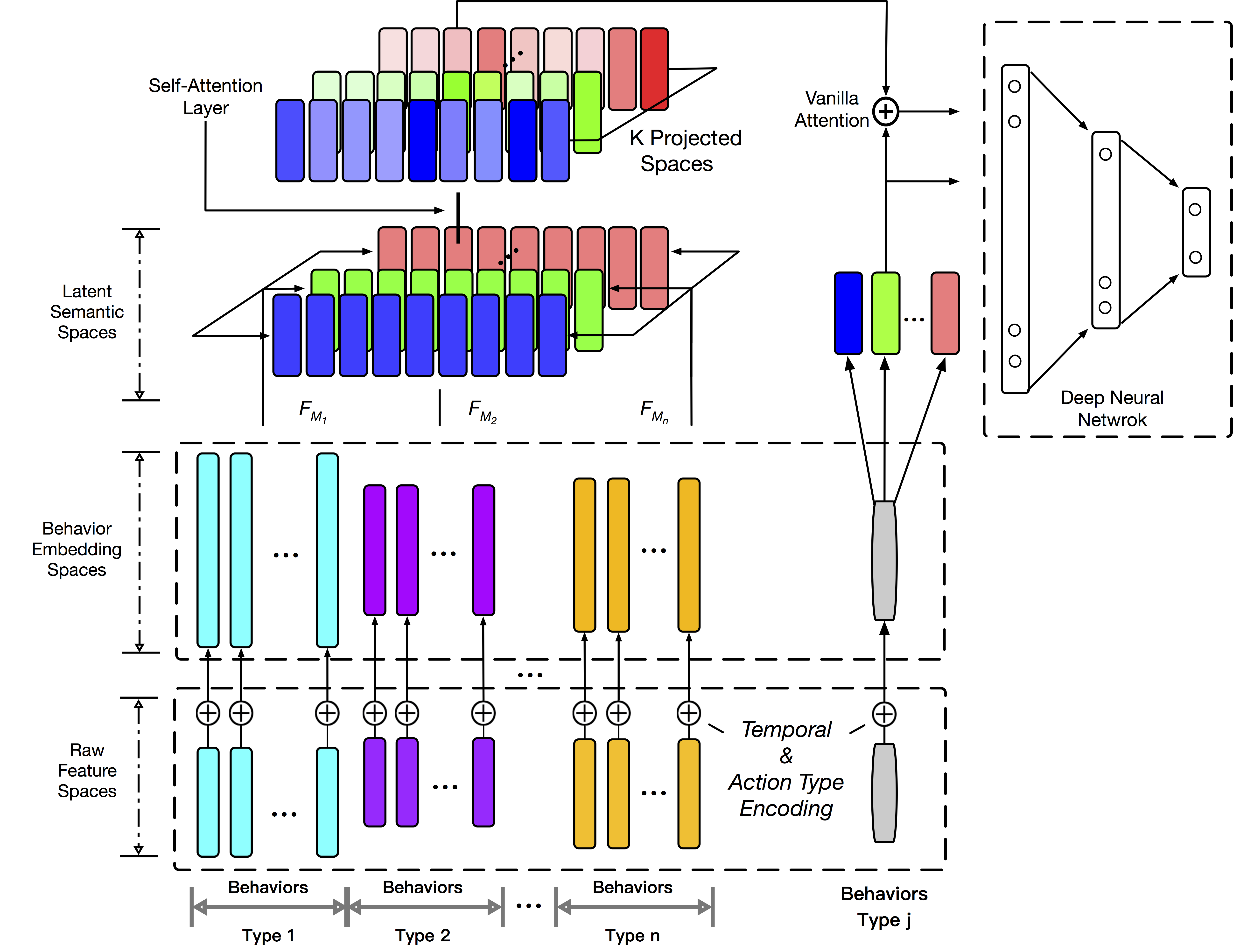}  
\caption{Attention-Based Heterogeneous Behaviors Modeling Framework} 
 \label{fig:model} 
\end{figure*}

This paper focuses on general user behaviors that can be interpreted using the binary relation between a user and an entity object. 
We formulate a user behavior as a tuple $\{a, o, t\}$, where $a$ stands for the behavior type describing the action a user takes, $o$ is the object that the behavior acts on, and $t$ is the timestamp when the behavior happens. Note that, while $a$ and $t$ are the atomic features representing the behavior type and action time, $o$ is represented as all its belonging features.
So that,  a user can be represented as all his/her behaviors $U=\{(a_j, o_j, t_j)|j=1,2,...,m\}$.

We illustrate the overall attention-based heterogeneous user behavior modeling framework in Figure \ref{fig:model}.
We divide the framework into several blocks, namely raw feature spaces, behavior embedding spaces, latent semantic spaces, behavior interaction layers and downstream application network. Next, we discuss all these blocks in detail.

\eat{We first define a user behavior sequence $U_{bg_c}=\{Tuple_i|i=1,2,...,k\}$, where $Tuple_i=\{bg_c, a_i, t_i, o_i\}$ with the meaning of behavior type, action type, event timestamp and target object features respectively. The raw feature sets of $o_i$ vary with the different bids.}

\subsection{Raw Feature Spaces}

We first partition the user behavior tuples  $U=\{ (a_j, o_j, t_j) |  j=1,2,...,m\}$ into different behavior groups $G=\{bg_1, bg_2, ..., bg_n\}$ according to the target object types, where  $bg_i \bigcap bg_j = \o $ and $U=\bigcup\limits_{i=1}^n{bg_i}$.
Within each $bg_i$, the raw feature spaces of the objects are the same, which may contain both the continuous features and the categorial features depending on how we model the target object.  Then we can use group-specific neural nets to build up the behavior embeddings.

\eat{Different user behaviors may have 
The model divides user behaviors in different groups according to the target object types in the behavior tuples. Within each group $bg_i$, different behaviors could have different raw features, either categorial or real valued.}

\subsection{Behavior Embedding Spaces}

In each behavior group $bg_i$, we embed the raw features of any behavior tuple  $(a_j, o_j, t_j)$ using the same embedding building block, 
$u_{ij} = f_i(a_j, o_j, t_j)$. The raw features of the target object $o_j$ are concatenated and fed into the feedforward neural network, while $a_j$ and $t_j$ are encoded separately that will be discussed below. 

{\sl Temporal \& Behavior Type Encoding.}
Each behavior may have different action time, which is a very important continuous feature, providing a good opportunity to cope with sequence data without the enrollment of any CNN or RNN-structures.
In order to preserve the temporal information, we use a temporal encoding methods similar as the positional embedding mentioned in \cite{cnn_fb}, with the embedding content to be the concatenation of both time and action type.
However, we found it very difficult in learning a good embedding directly on this continuous time feature using embedding concatenation or addition.
The better way we found is to bucketize the time feature into multiple granularities and perform categorial feature lookups. For example, 
in the recommend task over the amazon dataset, we slice the elasped time w.r.t the ranking time
into intervals whose gap length grow exponentially, e.g., we map the time in range $[0,1), [1, 2), [2, 4), ..., [2^k, 2^{k+1}), ...$
to categorial feature of 0, 1, 2, ..., k+1, .... Different behavior groups may have different granularities of time slicing.
Similar as the temporal encoding, we perform a direct lookup on the behavior type, trying to capture the impacts of different behavior types.
\eat{hypothesizing that different behavior type has quite different sensitivity towards time.
By a direct lookup on the discrete type.}

Then we can encode any user behavior $u_{ij} \in bg_i$  as  
\[u_{ij} =  emb_i(o_j)+ lookup_i^t(bucketize_i(t_j)) + lookup_i^a(a_j) \]
where $emb_i$ is the embedding concatenation for any object $o \in bg_i$. 
The outputs of the behavior embedding space are the list of vectors in all behavior groups,
\[B = \{u_{bg_1}, u_{bg_2},... ,u_{bg_n}\}\]
where $u_{bg_i}$ is the set of all behavior embeddings belonging to behavior group $i$, $u_{bg_i}=concat_j (u_{ij})$. Note that, the shape of the embeddings for each group  may be different, since
1) the numbers of behaviors in each group vary from user to user, 2) the information carried for each type of behavior could be in different volume. E.g., an item behavior may contain  much more information than a keyword search behavior, since the item may reveal user preference towards brand, price, style, tastes, etc, while one short query keyword can only show limited intents of a user.

{\sl Embedding Sharing.} We enable the embedding sharing when multiple groups of objects have features in common, \eat{There may also have inter-group embedding-sharing for the same type of raw feature,} e.g, shop id, category id that can be both shared by objects like items and coupons in an e-business recommendation system, referring to the same entities. Note that we don't share the temporal encoding lookups across behavior groups, since the time impact for each type of behavior may vary a lot. For example, the time impact is dominating on the coupon receiving behavior, since it's a strong short-term intention, while buying shoes of a certain brand may be of long-term interest, where the temporal information could be less important. 

\eat{We can benefit from the embedding sharing by several aspects. First, embedding sharing can dramatically reduce parameters that the model needs to learn. Second,  it's necessary if we need to cope with same entities in different behavior types, and the sparsity issue can also be reduced by sharing embedding matrix. 
}

\subsection{Latent Semantic Spaces}

Lots of works have shown that linear projections over multi-subspaces could improve the performance in various tasks. We argue that heterogeneous behaviors could have very different expressive power, thus their embedding spaces could be in both different sizes and meanings. Linear projection here plays a role to put them into the same semantic space, where connections or comparisons can be made. So we map each behavior into $K$ latent semantic spaces. We explain this idea in Figure \ref{fig:model} that, each behavior group is represented as a composite color bar, which is then factorized into RGB colors where further operations can be taken within the comparable atomic color spaces.

Given the user behaviors' representation $B$,
\eat{\[S = relu(concat^{(0)}(u_{bg_1} M_{1}, u_{bg_2} M_{2}, ..., u_{bg_n} M_{n}))\]}
we first project the variable-length behaviors in different groups into fix-length encoding vectors
\begin{equation}
S = concat^{(0)}( \mathcal{F}_{M_1}(u_{bg_1}), \mathcal{F}_{M_2}(u_{bg_2}), ..., \mathcal{F}_{M_n}(u_{bg_n}))
\end{equation}
where the symbol $\mathcal{F}_{M_i}$ represents the projection function which is parameterized by $M_i$.
$\mathcal{F}_{M_i}$ maps a behavior in group $i$ onto the overall space of dimension size $s_{all}$.
Then the projected behavior embeddings in each spaces are
\begin{equation}
S_k= \mathcal{F}_{P_k}(S)
\end{equation}
where $\mathcal{F}_{P_k}$ is a projection function for the $k\mbox{-}th$ semantic space. In this paper, all the projection function $\mathcal{F}_\theta$ is set to be a single layer perceptron parametrized by $\theta$ with $relu$ as the activation function.

Let the number of all behaviors of a user be $n_{all}$, the number of behaviors in group $i$ be $n_i$, the dimension of the feature embedding space for that group be $b_i$, the dimension of the $k\mbox{-}th$ semantic space be $s_k$.
Then $u_{bg_i}$ is in shape of $n_i\mbox{-}by\mbox{-}b_i$,
$S_k$ is in shape of $n_{all}\mbox{-}by\mbox{-}s_k$.
The  superscript on $concat$ function indicates which dimension the concatenation is performed on.

\subsection{Self-Attention Layer}

This layer tries to capture the inner-relationships among each semantic space. The intense of each behavior could be affected by others, such that attentions for some behaviors could be distracted and the others could be strengthened. 
This is done by the self-attention mechanism. The outputs of this layer could be regarded as the behavior representation sequence taking considerations of the impacts of the others in each latent space.
\eat{showing that a user is represented by all the things it does, with a strong or weak attention.}

We exploit similar self-attention structure mentioned in \cite{google_attention} for Machine Translation task, with some customized settings.
We calculate each attention score matrix $A_k$ in the $k\mbox{-}th$ semantic space as

\begin{equation}
A_k = softmax(a(S_k, S; \theta_k))
\label{eq:att_score}
\end{equation}
where each row of $A_k$ is the score vector w.r.t all the attention vectors in that subspace, the softmax function guarantees that all scores for a behavior sums to one. 
The score function $a(S_k, S; \theta_k)$ measures the impacts among all behaviors in the $k\mbox{-}th$ semantic space, and in this paper we choose the bilinear scoring function as in \cite{att_score}  
\begin{equation}
a(S_k, S; \theta_k)=S_kW_kS^T
\end{equation}
Then the attention vectors of space $k$ are
\begin{equation}
C_k = A_k \mathcal{F}_{Q_k}(S) 
\end{equation}
where $\mathcal{F}_{Q_k}$ is another projection function that maps $S$ onto the $k\mbox{-}th$ semantic space, which in this paper is a single layer perceptron with $relu$ activation fucntion.
\eat{  parameter which helps to rebuild the attention vectors}
Then $C_k$ is in shape of $n_{all}\mbox{-}by\mbox{-}s_k$.

\eat{We can see that each output vector is the combination of all the behavior vectors considered within the $r$-th order relationships, where $r$ is the current level of all the stacked attention layers.  In order to introduce higher-order interactions among the user behaviors, multiple self-attention layers could be added. }

These vectors from different subspaces are concatenated together and then reorganized by
\begin{equation}
C = \mathscr{F}_{self}(concat^{(1)}(C_1, C_2, ..., C_K))
\end{equation}
where $\mathscr{F}$ is a feedforward network with one hidden layer,
aiming to provide non-linear transformation over each behavior vector after attention-based pooling.
The output of $\mathscr{F}$ keeps the same shape as the input, and we successively perform drop out, residual connections and layer normalization on the outputs.

\eat{\[\mathscr{F}(x) = relu(xW_1+b_1)W_2+b_2\] }

\subsection{Downstream Application Network}

With the generated user behavior models, we can ensemble various kinds of neural networks according to the downstream task requirement. In this paper, we focus on evaluating the recommendation tasks, and we set the downstream application network to be a point-wise or a pair-wise fully connected neural network. 

{\sl Vanilla Attention. } For both the point-wise and the pair-wise model, a vanilla attention is performed to produce the final context vector $e_u^t$ for user $u$ w.r.t. the embedding vector $q_t$ to be predicted. This  follows the similar procedure in the self-attention phase,
\eat{except that the first $S_i$ in equation \ref{eq:att_score} is now a vector, representing the ranking behavior embedding which is projected on the $i\mbox{-}th$ space.}
\begin{equation}
\begin{aligned}
\vec{h_t} & = \mathcal{F}_{M_{g(t)}}(\vec{q_t}), \hspace{0.4cm} \vec{s_k} =  \mathcal{F}_{P_k}(\vec{h_t}) \\
\vec{c_k} & = softmax(a(\vec{s_k}, C; \theta_k)) \mathcal{F}_{Q_k}(C) \\
\vec{e_u^t} & = \mathscr{F}_{vanilla}(concat^{(1)}((\vec{c_1}, \vec{c_2}, ..., \vec{c_K})))
\end{aligned}
\label{eq:v_att}
\end{equation}
\eat{\begin{equation}
\label{eq:v_att}
\end{equation}}
where $g(t)$ maps behavior $t$ to its behavior group number, and $\mathscr{F}_{vanilla}$ has different parameters with $\mathscr{F}_{self}$.

For a point-wise prediction model, the probability of each behavior $t$ that will be taken is predicted. We extract the raw feature of $t$, map these features into the behavior embedding space according to the building block of its behavior group $q_t$. Then we perform a vanilla attention as shown above. 
The final loss function is the sigmoid cross entropy loss
\begin{equation}
- \sum_{t,u}{y_t \log{\sigma(f(h_t,e_u^t))} + (1-y_t )\log{(1-\sigma(f(h_t,e_u^t)))}}
\label{eq:loss_pointwise}
\end{equation}
where $f$ is a ranking function whose input is the ranking behavior embedding $q_t$ and the user encoding  $e_u^t$  w.r.t behavior $t$ computed by vanilla-attention.
$f$ can be either a dot-product function or a more complexed deep neural network. 

It is easy to rewrite the downstream network to support pair-wise training and evaluation, and we omit here due to limited space.

\eat{
For a pair-wise prediction model, a set of partial order is defined as
\begin{equation}
D_S = \{(u,i,j)|i \in I_u^+ \land j \in I \backslash I_u^+\} 
\end{equation}
where $I$ is the set of all behaviors and $I_u^+$ is the behavior subset that user $u$ has performed.
We compute the embedding difference of behavior $i$ and $j$ in each subspace as $\Delta=\{h_i-  h_j\}$, and then replace the term $h_t$ with 
$\Delta$ in equation \ref{eq:v_att} to calculate $e_u^{i>j}$.
The loss function in pair-wise prediction model is defined as

\begin{equation}
L = - \sum_{(u,i,j) \in D_S}{ \log{\sigma(f(\Delta, e_u^{i>j}))}}
\end{equation}
}
\eat{Note that, the user behaviors above need not to be restricted to single type or even group, meaning that we can build a unified model to perform multi-task that predicts different
types of behaviors at the same time.
}

\eat{This neural network can be of any structure according to the downstream task requirement.
In this paper, we focus on evaluating the recommendation tasks, and we set the downstream application network to be a point-wise fully connected neural network for simplicity. 
}
\eat{either a ranking model or a generation model, w.r.t different task demands.}
 
\section{Experiment}

In this section, we first evaluate our method in context-aware recommendation tasks with only one type of behavior, trying to show the benefits of using self-attention to encode user behaviors. Then we collect a multi-behavior dataset from an online recommendation service, and evaluate how the heterogeneous behavior model performs in recommendation tasks.
We further explore the model to support multi-task using one unified model, which can predict all kinds of behaviors at the same time.

\subsection{Dataset}

Amazon Dataset.
We collect several subsets of amazon product data as in \cite{amz_mc}, which have already been reduced to satisfy the 5-core property, such that each of the remaining users and items have 5 reviews each \footnote{http://jmcauley.ucsd.edu/data/amazon/}.
The feature we use contains user id, item id, cate id and timestamps.
The statistics of the dataset is show in table \ref{table:DataSets}. Let all the user behaviors for $u$ be $(b_1, b_2, ..., b_k, ..., b_n)$, we make the first k behaviors of $u$ to predict the $k\mbox{+}1\mbox{-}th$ behavior in the train set, where $k=1, 2, ..., n\mbox{-}2$, and we use the first $n\mbox{-1}$ behaviors to predict the last one in the test set.

\begin{table}[htp]
\begin{center}
\begin{tabular} { |c|c|c|c|c| } \hline %
Dataset & $\#$ Users & $\#$ Items  & $\#$ Cates & $\#$ Samples \\ \hline \hline

{\sl Electro.} & 192,403 & 63,001 & 801 & 1,689,188\\ \hline

{\sl Clothing.} & 39,387 & 23,033 & 484 & 278,677\\ \hline

\end{tabular}
\caption {Statistics of Amazon DataSets}
\label{table:DataSets}
\end{center}
\end{table}

Taobao Dataset. We collect various kinds of user behaviors in a commercial e-business website \footnote{http://www.taobao.com}. 
We extract three groups of behaviors, namely item behavior group, search behavior group and coupon receiving behavior group.
Item behaviors list all the user action types on items, such as browse, mark, buy, etc. The feature set of an item contains item id, shop id, brand id, category id.
Search behavior is expressed as the search tokens, and has only one type of action. 
In coupon receiving behavior group, coupon features are consist of coupon id, shop id and coupon type. This group also has only one action type.
All above behaviors have a time feature that represent the timestamp when that behavior happens.
The statics of each categorial is shown in table \ref{table:ali_data}. This dataset is guaranteed that each user has at least 3 behaviors in each behavior group.

\begin{table*}[htp]
\begin{center}
\begin{tabular} { |c|c|c|c|c|c|c|c|c|c| } \hline %

Dataset & $\#$Users & $\#$Items  & $\#$Cates & $\#$Shops &  $\#$Brands & $\#$Queries & $\#$Coupons & $\#$Records &Avg Length \\ \hline \hline

{\sl Multi-} & \multirow{2}{*}{30,358} & \multirow{2}{*}{447,878} & \multirow{2}{*}{4,704} & \multirow{2}{*}{109,665} & \multirow{2}{*}{49,859} & \multirow{2}{*}{111,646} & \multirow{2}{*}{64,388} & \multirow{2}{*}{247,313} & \multirow{2}{*}{19.8} \\
{\sl Behavior.} & & & & & & & & & \\ \hline

\end{tabular}
\caption {Statistics of Multi-Behavior DataSets}
\label{table:ali_data}
\end{center}
\end{table*}

\subsection{Competitors}

\begin{itemize}
\item BPR-MF: Bayesian Personalized Ranking \cite{bpr} is a pairwise ranking framework.
It trains on pairs of a user's positive and negative examples, maximizing the posterior probability of the difference given the same user.
The item vector is a concatenation of item embedding and category embedding, each of which has a dimension of 64, the user vector has a dimension of 128.

\eat{
\item BPR: Bayesian Personalized Ranking \cite{bpr} is a pairwise ranking framework for implicit feedback data. Let the user vector be $u$ and item vector be $i$,
It trains on pairs of a user's positive and negative examples, 
maximizing the probability of the difference given the same user
\[ L = - \sum_{(u,i,j) \in D_S}{ \log \sigma( {\vec{u}\cdot(\vec{i}-\vec{j})}} ) + \lambda_\Theta || \Theta ||^2 \]
where $\sigma(\cdot)$ is  the logistic sigmoid function, and $\Theta$ is the model parameter vector.
The item vector is a concatenation of item embedding and category embedding, each of which has a dimension of 64, the user vector has a dimension of 128.
}
\item Bi-LSTM: We implement a Bi-LSTM method to encode the user behavior history,
whose difference with \cite{seq_click} is that we use a bidirectional LSTM as the encoder because of better performance in our experiments.
Then we concatenate the two final hidden states to the user embedding.
The stacked LSTM depth is set to be 1.

\item Bi-LSTM+Attention: We add a vanilla attention on top of the Bi-LSTM method mentioned above.

\item CNN+Pooling: We use a CNN structure with max pooling to encode the user history as in \cite{cnn_wsdm,cnn_nlp}. 
We use ten kinds of window sizes from one to ten for extracting different features, and all of the feature map have the same kernel size with 32.
Then we apply a max-over-time pooling operation over the feature map, and pass all pooled features from different maps to a fully connected layer to produce final user embedding.
\end{itemize} 

\begin{figure}
\centering
\includegraphics [width=3.2 in, height=2.2 in]{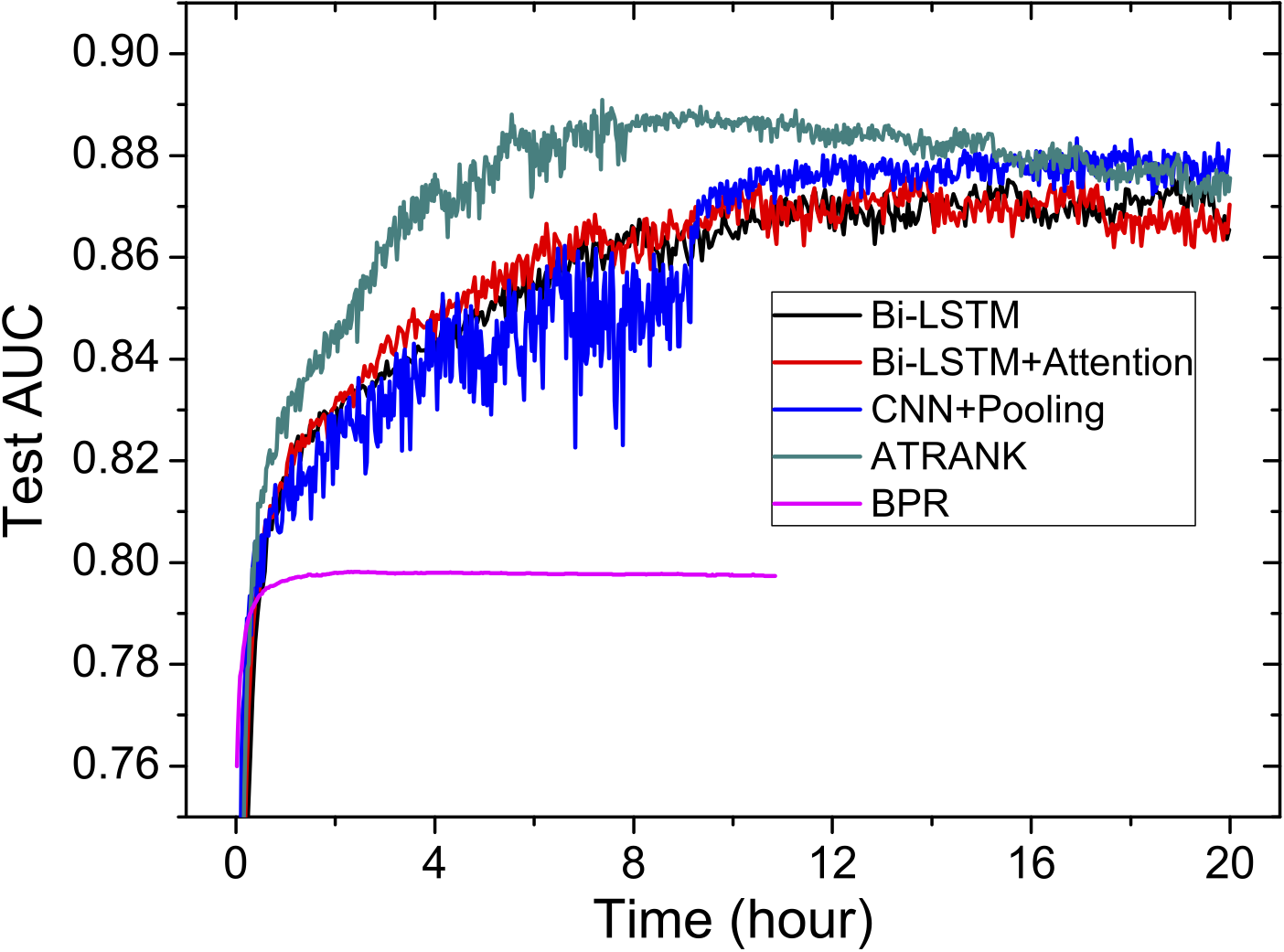}  
\caption{AUC Progress in Amazon Electro Test Set} 
 \label{fig:progress} 
\end{figure}

\subsection{Hyperparameters}

Our method and all competitors use the common hyperparameters as follows.

\begin{itemize}
\item Network Shape. We set the dimension size of each categorial feature embedding to be 64, and we concat these embeddings as the behavior representation in the behavior embedding space. The hidden size of all layers is set to be 128. The ranking function $f$ is simply set to be the dot product in these tasks. As we observe better performance with point-wise ranking model, we omit the results of pair-wise models here for simplicity. 
For ATRank, we set the number of the latent semantic spaces to be 8, whose dimension sizes sum to be the same as the size of the hidden layer.
\eat{which doesn't increase the amount of network parameters.}
\item Batch Size. The batch size is set to be 32 for all methods.
\item Regularization. The l2-loss weight is set to be 5e-5.
\item Optimizer. We use SGD as the optimizer and apply exponential decay which learning rate starts at 1.0 and decay rate is set to 0.1.
\eat{
\item Optimizer. We use SGD as the optimizer and an adaptive learning rate which starts at 1.0.}
\end{itemize}

\subsection{Evaluation Metrics}

We evaluate the average user AUC as following:
\[AUC=\frac{1}{|U^{Test}|}\sum_{u\in U^{Test}}\frac{1}{|I_u^+||I_u^-|}\sum_{i\in I_u^+}\sum_{j\in I_u^-}\delta(\hat{p_{u,i}} > \hat{p_{u,j}})\]
where $\hat{p_{u,i}}$ is the predicted probability that a user $u\in U^{Test}$ may act on $i$ in the test set and $\delta(\cdot)$ is the indicator function.

\subsection{Results on Single-type Behavior Dataset}

We first illustrate the average user AUC of all the methods for the amazon dataset in table \ref{table:auc_amz}.
We can see that ATRank performs better than the competitors especially when the user behavior is dense,
which reveals the benefits of the self-attention based user behavior modeling. 

Table \ref{table:avg_score_time} illustrates the average vanilla-attention score for different time buckets over the whole amaon dataset, which can be inferred that time encoding via self-attention can also incorporate the time information as a replacement of RNN/CNN.

\eat{This is because our network has more parameters w.r.t the competitors and it's much harder to learn when the data is sparse.}

Then we show how the average user AUC for the test set evolves along with the training procedure in Figure \ref{fig:progress}. We can see that 
ATRank converges much faster than RNN or CNN-based methods in training, since ATRank does not incur less-parallelizable operations
as in RNN, and any behavior could be affected by the other in just one layer of self attention. \eat{ which leads to less computational dependency and better performance. }
Though BPR is faster in training, it has a very pool performance.

\subsection{Results on Multi-type Behavior Dataset}

\eat{The multi-behavior tasks are to predict each type of behaviors using all types of user behaviors.
In the multi-type dataset we build, we try to use the behaviors to predict while the user behaviors are extracted not limited to that scenario.
So the training sample of the item click behavior is the online click logs, 
which contains both positive and negative samples. 
We leave the last page view of each user in the test set and the rest of the }

We train the general model to evaluate the probability of the next user action, including all types of behaviors.
We evaluate 7 tasks regarding different parts of the dataset as the train set and the test set.

{\sl One2one model. }
First we train three models using only one type of the behaviors and predict on the same type of behavior, namely item2item, coupon2coupon, query2query as baseline, we refer these type of model as one2one model.

{\sl All2one model.}
Then we train another three models using all types of the behaviors as user history and predict each type of behavior separately, namely all2item, all2coupon, all2query, we refer these models as all2one model.

{\sl All2all model.}
Finally we train a unified model for multi-task evaluation, using all types of user behaviors to predict any type of behavior at the same time, and we refer this model as all2all model.

{\sl Sample generation.}
In the multi-type behavior dataset we build, the item samples are the click actions in one of the inner-shop recommendation scenarios, which are collected from the online click log of that scenario containing both positive and negative samples. The user behaviors, however, are not limited to that scenario. We build the coupon samples by using the behaviors before the coupon receiving action, and the negative sample is randomly sampled in the rest of the coupons. The query samples are made similar with the coupon samples.

\eat{We train the general model to evaluate the probability of the next user action, including all types of behaviors.}

Then we show the average user AUC of all three models in table \ref{table:one2one} respecitvely.
We can see that, all2one model utilizes all types of user behaviors so that it performs better than the one2one models which only use the same type of behavior.
The all2all model actually performs three different tasks using only one model, 
which shows that it can benefit from these mix-typed training process and achieve comparable performance with the highly optimized individual models.

\begin{table}[htp]
\begin{center}
\begin{tabular} { |c|c|c| } \hline %
Dataset & Electro.  & Clothe.  \\ \hline \hline

{\sl BPR} & 0.7982 & 0.7061 \\ \hline

{\sl Bi-LSTM} &  0.8757 & 0.7869 \\ \hline

{\sl Bi-LSTM + Attention} & 0.8769 & 0.7835 \\ \hline

{\sl CNN + Max Pooling}  & 0.8804 & 0.7786 \\ \hline

{\sl ATRank} & \textbf{0.8921} & \textbf{0.7905}\\ \hline

\end{tabular}
\caption {AUC on Amazon Dataset}
\label{table:auc_amz}
\end{center}
\end{table}

\begin{table*}[htp]
\begin{center}
\begin{tabular} { |c|c|c|c|c|c|c|c| } \hline %
{\sl Time Range(\#Day)} &  [0, 2)  & [2, 4)  & [4, 8) & [8, 16) & [16, 32) & [32, 64) & [64, 128) \\ \hline
{\sl Avg Att-Score} & 0.2494 & 0.1655 & 0.1737& 0.1770 & 0.1584 & 0.1259 & 0.1188 \\ \hline
\end{tabular}
\caption {Average Attention Score for Different Time Bucket over Amazon Dataset}
\label{table:avg_score_time}
\end{center}
\end{table*}

\begin{table}[htp]
\begin{center}
\begin{tabular} { |c|c|c|c|} \hline %
Predict Target & Item & Query & Coupon  \\ \hline \hline

{\sl Bi-LSTM} &  0.6779 & 0.6019 & 0.8500 \\ \hline
{\sl Bi-LSTM + Attention} & 0.6754 & 0.5999 & 0.8413 \\ \hline
{\sl CNN + Max Pooling}  & 0.6762 & 0.6100 & 0.8611 \\ \hline
{\sl ATRank-one2one} &  0.6785 & 0.6132 & 0.8601 \\ \hline
{\sl ATRank-all2one} &  \textbf{0.6825} & \textbf{0.6297} & \textbf{0.8725} \\ \hline
{\sl ATRank-all2all} &  0.6759 & 0.6199 & 0.8587 \\ \hline

\end{tabular}
\caption {AUC on Ali Multi Behavior Dataset}
\label{table:one2one}
\end{center}
\end{table}

\eat{
\begin{table}[htp]
\begin{center}
\begin{tabular} { |c|c|c|c|} \hline %
Predict Target & Item & Coupon & Query  \\ \hline \hline

{\sl ATRank} &  0.6825 & 0.6084 & 0.8664 \\ \hline

\end{tabular}
\caption {AUC on Multi Behavior Dataset: All2one Task}
\label{table:all2all}
\end{center}
\end{table}

\begin{table}[htp]
\begin{center}
\begin{tabular} { |c|c|c|c|} \hline %
Predict Target & Item & Coupon & Query  \\ \hline \hline

{\sl ATRank} &  0.6711 & 0.6134 & 0.8490 \\ \hline

\end{tabular}
\caption {AUC on Multi Behavior Dataset: All2all Task}
\label{table:all2one}
\end{center}
\end{table}
}

\subsection{Case Study}

\begin{figure}[htp]
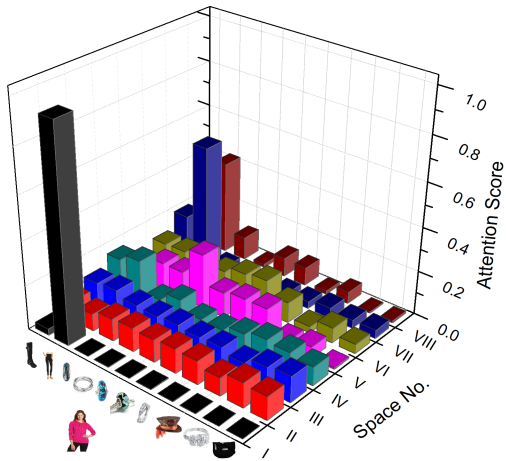

\centering
 
\subfigure[Average Vanilla-Attention Score over All Latent Space]{
       
       \psfig{figure=case_v_att_avg.png,width=3.2 in,height=1.2 in}
         \label{fig:case_v_att_avg}
      }
\subfigure[Vanilla-Attention Score in Different Latent Spaces] {
        \psfig{figure=case_v_att.png,width=3.2 in,height=2.4 in}
        \label{fig:case_v_att_space}
      }

\caption{Case Study on Vanilla-Attention in ATRank} 
 \label{fig:case_v_att} 
\end{figure}

We try to explain the effects of self-attention and vanilla-attention in our model separately by case study.
We visualize a user buying sequence from amazon dataset in Figure \ref{fig:case_v_att_avg}, which are one bag for woman, one ring, one hat for men, five jewelry of all kinds, one women's legging and women's shows respectively, the ranking item is a women's hoodie. 

\begin{figure*}
\centering
\includegraphics [width=6.4 in, height=2.7 in]{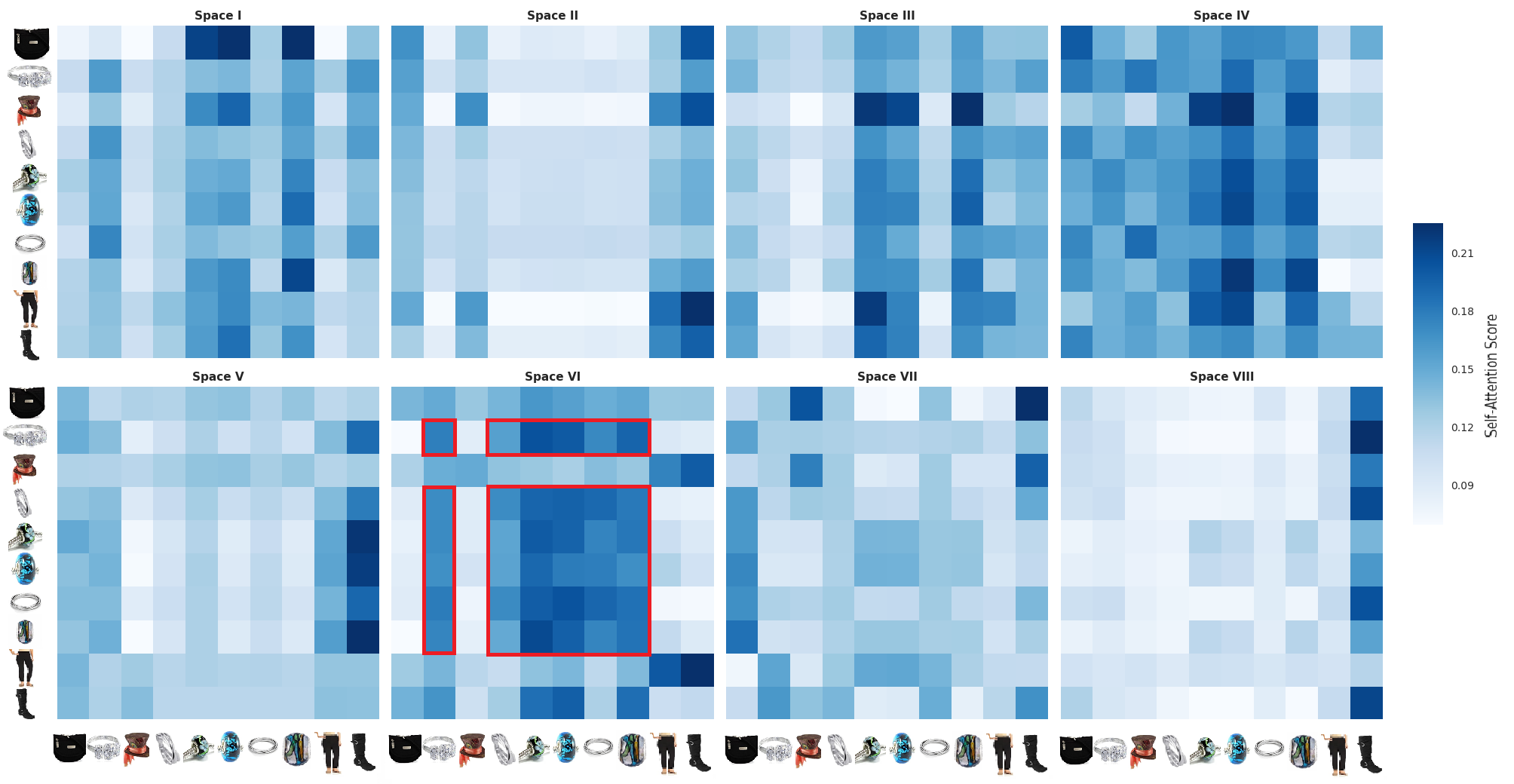}  
\caption{Case Study: Heatmap of Self-Attention in ATRank} 
 \label{fig:case_self_att} 
\end{figure*}

\eat{
\begin{figure}
\centering
\includegraphics [width=3.2 in, height=1.4 in]{all_new.png}  
\caption{Case Study on Vanilla-Attention in ATRank} 
 \label{fig:case_v_att} 
\end{figure}
}

\eat{
\begin{figure}
\centering
\includegraphics [width=3.2 in, height=1.2 in]{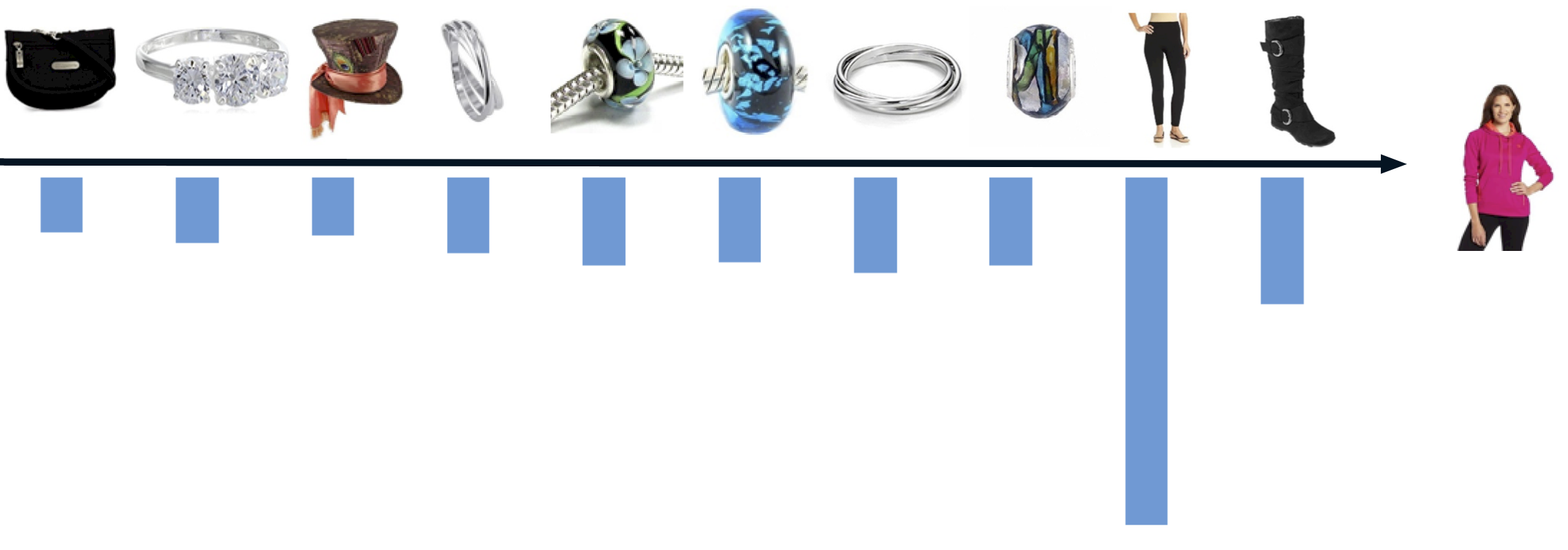}  
\caption{1.jpg}
\includegraphics [width=3.2 in, height=2.4 in]{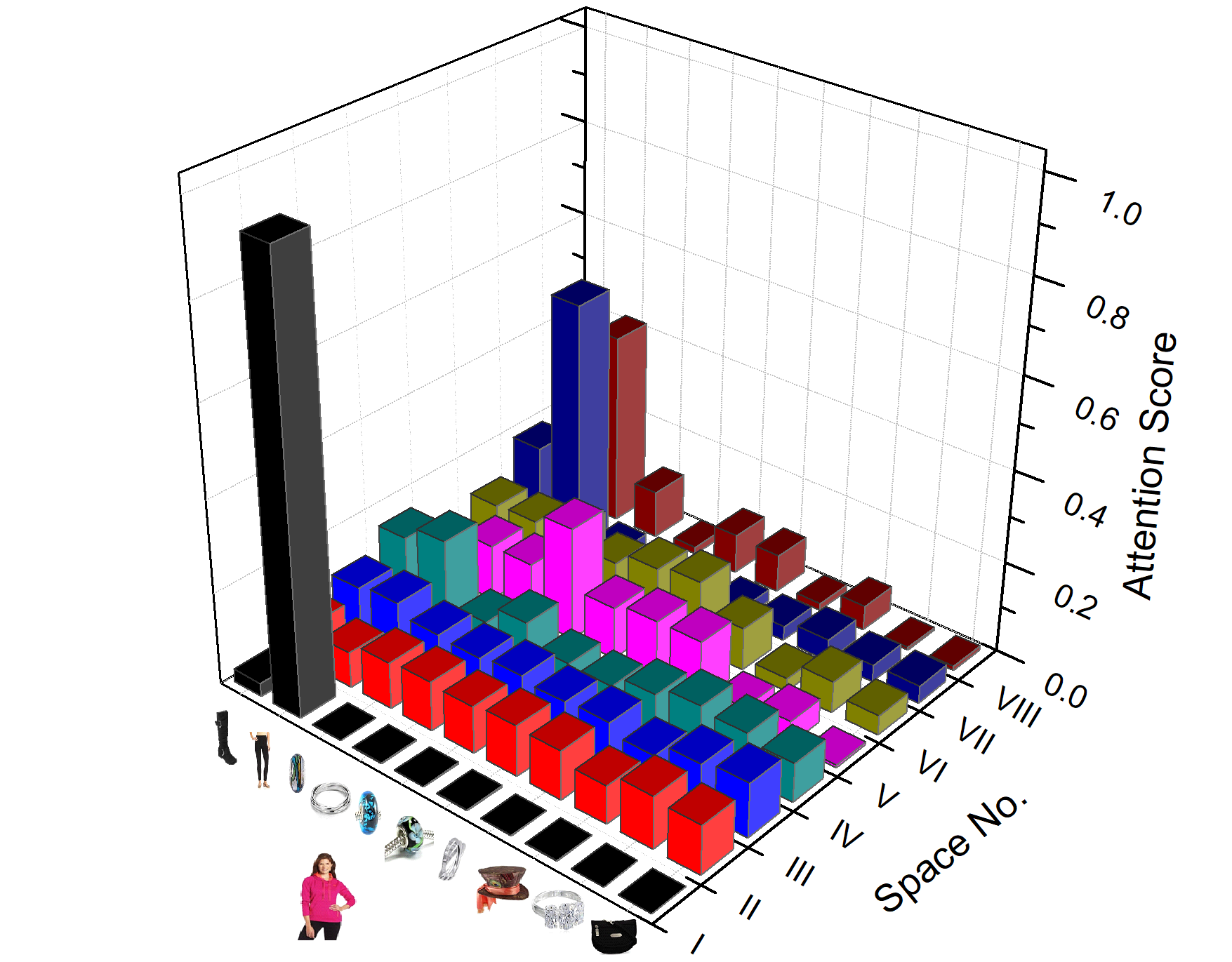}  

\caption{Case Study on Vanilla-Attention in ATRank} 
 \label{fig:case_v_att} 
\end{figure}
}

We first illustrate how the self-attention scores look like among those behaviors in different semantic spaces in Figure \ref{fig:case_self_att}.
Here the items that a user has bought form a timeline placed repeatedly as coordinate, and the heat map matrix represent the self-attention scores between any two behaviors in several different semantic space. Note that, the score matrix is normalized through row-oriented softmax so that it is asymmetric. 

Then we have several observations from the Figure \ref{fig:case_self_att}. 
First, different semantic space may focus on different behaviors, since the heat distributions vary a lot in all the 8 latent spaces.
Second, in some spaces, e.g., space \uppercase\expandafter{\romannumeral1}, \uppercase\expandafter{\romannumeral2}, \uppercase\expandafter{\romannumeral3} and \uppercase\expandafter{\romannumeral8}, the relative trends of the attention score vectors keep the same for each row's behavior, while the strength is varied. The higher scores tend to gather around some specific columns. These spaces may consider more about the overall importance of a behavior among them.
Third, in some other spaces, the higher scores tend to form a dense square. Take the space \uppercase\expandafter{\romannumeral6} as an example, behavior 2, 4, 5, 6, 7, 8 may form a group of jewelries, among which user behaviors have very high inner attention scores.
Those spaces probably consider more about relationships of category similarities. It then can be concluded that self-attention models the impacts among the user behaviors in different semantic spaces. 

\eat{
while which a few items can always dominate the attention score in each row when the item are similar, meaning that it may have an effect of aggregation on very similar items. 3) Items in the same categories can result .
}
Then we study the effects of vanilla-attention in ATRank in Figure \ref{fig:case_v_att}. 
\eat{The ranking item is a women's hoodie shown in the right-most place. }
We can see that the attention score for each latent space also varies a lot in Figure \ref{fig:case_v_att_space}. Some latent spaces, e.g., space \uppercase\expandafter{\romannumeral1}, \uppercase\expandafter{\romannumeral7}, \uppercase\expandafter{\romannumeral8}, highlight only the most correlated behaviors, while others like \uppercase\expandafter{\romannumeral2}, \uppercase\expandafter{\romannumeral3} show the average aggregation of all the behaviors. Note that, the space number of vanilla attention does not have a necessarily correspondence with that of self-attention, because there is a feedforward neural network that performs another non-linear projections on the concatenated self-attention vectors.

\eat{
The score heat map in this figure is a 8-by-10 matrix that each row represents a normalized attention score in a semantic space. We can infer that, 1) the attention scores usually grow as the behavior time approaches the target time. 2) The attention scores enable us to capture the most related items precisely. 3) different semantic space has  considerations on different aspects. 
}

\eat{
{\sl Self attention.} Self attention

{\sl Vanilla attention.}

1.2. 

2. Heterogeneous Modeling
1. queries, coupons, items.
}
\section{Conclusion}
This paper proposes an attention-based behavior modeling framework called ATRank, which is evaluated on recommendation tasks. ATRank can model with heterogeneous user behaviors using only the attention model. Behaviors interactions are captured using self-attention in multiple semantic spaces. The model can also perform multi-task that predict all types of user actions using one unified model, which shows comparable performance with the highly optimized individual models. In the experiment, we show that our method achieves faster convergence while obtains better performance. We also give a case study that shows the insight of the how attention works in ATRank.

\eat{This model can also predict the user's behaviors of different types. 

Some interesting points can be considered in the future work.
How to add penalty terms to make multiple semantic spaces produce different semantics? 
As the information volume of each semantic space may be different, will variable sized semantic space help to improve the performance?
The internal behavior of self-attention needs to be studied.
   }
   
\section*{Acknowledgments}
This work was partially supported by National Key Research and Development
Program No. 2016YFB1000700, NSFC under Grant No.
61572040, and Alibaba-PKU joint Program.

\bibliography{reference}
\bibliographystyle{aaai}
\end{document}